\documentclass[10pt, a4paper]{article}
\usepackage{lrec}
\usepackage{graphicx}
\usepackage{tabularx}
\usepackage{multirow}
\usepackage{multicol}
\usepackage{soul}
\usepackage{microtype}
\usepackage{todonotes}
\usepackage{stfloats}

\usepackage{epstopdf}
\usepackage[latin1]{inputenc}

\usepackage{hyperref}
\usepackage{xstring}
\usepackage{fontawesome}

\newcommand{\secref}[1]{\StrSubstitute{\getrefnumber{#1}}{.}{}}

\newcommand{\cfbox}[2]{%
    #2
}

\title{AMALGUM -- A Free, Balanced, Multilayer English Web Corpus}

\name{
    Luke Gessler,
    Siyao Peng,
    Yang Liu,
    Yilun Zhu,
    Shabnam Behzad,
    Amir Zeldes
}

\address{
    Corpling Lab \\
    Georgetown University \\
    \{lg876, 
    sp1184, 
    yl879, 
    yz565, 
    sb1796,
    az364\}@georgetown.edu\\
}

\abstract{We present a freely available, genre-balanced English web corpus totaling 4M tokens and featuring a large number of high-quality automatic annotation layers, including dependency trees, non-named entity annotations, coreference resolution, and discourse trees in Rhetorical Structure Theory. By tapping open online data sources the corpus is meant to offer a more sizable alternative to smaller manually created annotated data sets, while avoiding pitfalls such as imbalanced or unknown composition, licensing problems, and low-quality natural language processing. We harness knowledge from multiple annotation layers in order to achieve a ``better than NLP'' benchmark and evaluate the accuracy of the resulting resource.
  \\ 
\newline  }

\begin{document}

\maketitleabstract

\section{Introduction}\label{sec:intro}

Corpus development for academic research and practical NLP applications often meets with a dilemma: development of high quality data sets with rich annotation schemes (e.g.~MASC \cite{IdeEtAl2010}) requires substantial manual curation and analysis effort and does not scale up easily; on the other hand, easy ``opportunistic'' data collection \cite{Kennedy1998}, where as much available material is collected from the Internet as possible, usually means little or no manual annotation or inspection of corpus contents for genre balance or other distributional properties, and at most the addition of relatively reliable automatic annotations such as part of speech tags and lemmatization (e.g.~the WaCky corpora \cite{BaroniBernardiniFerraresiEtAl2009}), or sometimes out-of-the-box, automatic dependency parsing (see the COW corpora, \cite{Schaefer2015}). This divide leads to a number of problems:

\begin{enumerate}
    \item The study of systematic genre variation is limited to surface textual properties and cannot use complex annotations, unless it is confined to very limited data
    \item For large resources, NLP quality is low since opportunistically-acquired content often deviates substantially from the language found in training data used by out-of-the-box tools
    \item Large open data sets for training NLP tools on complex phenomena (e.g.~coreference resolution, discourse parsing) are unavailable\footnote{The largest datasets for these tasks (e.g.~RST Discourse Treebank \cite{CarlsonEtAl2003}, OntoNotes \cite{WeischedelPradhanRamshawEtAl2012}), contain between 200K and 1.6 million words and are only available for purchase from the LDC.}
\end{enumerate}

In this paper, we attempt to walk a middle path, combining some of the best features of corpus data harvested from the web -- size, open licenses, lexical diversity -- and data collected using a carefully defined sampling frame (cf.~\newcite{Hundt2008}), which allows for more interpretable inferences as well as the application of tools specially prepared for the target domain(s). In addition, we focus on making discourse-level annotations available, including complete nested entity recognition, coreference resolution, and discourse parsing, while striving for accuracy that is as high as possible. Some of the applications we envision for this corpus include:

\begin{itemize}
    \itemsep0em 
    \item Corpus linguistic studies on genre variation and discourse using detailed annotation layers
    \item Active learning -- by permuting different subsets of the automatically annotated corpus as training data, we can evaluate performance on gold standard development data and automatically select those documents whose annotations improve performance on a given task 
    \item Cross-tagger validation -- we can select the subset of sentences which are tagged or parsed identically by multiple taggers/models trained on different datasets and assume that these are highly likely to be correct (cf.~\newcite{DerczynskiRitterClarkEtAl2012}), or use other similar approaches such as tri-training (cf.~\newcite{zhou2005tri})
    \item Human-in-the-loop/crowdsourcing -- outputting sentences with low NLP tool confidence for expert correction or crowd worker annotations, or using a hybrid annotation setup in which several models are used, and the tags that the models agree on are only reviewed by annotators and the rest are annotated from scratch \cite{berzak-etal-2016-anchoring}
    \item Pre-training -- the error-prone but large automatically produced corpus can be used for pre-training a model which can later be fine-tuned on a smaller gold dataset
\end{itemize}

Testing these applications is outside the scope of the current paper, though several papers suggest that large scale `silver quality' data can be better for training tools than smaller gold standard resources for POS tagging \cite{SchulzKetschik2020}, parsing \cite{Toutanova2005} using information from parse banks \cite{CharniakBlahetaGeEtAl2000}, discourse relation classification \cite{marcu-echihabi-2002-unsupervised}, and other tasks. Moreover, due to its genre diversity, this corpus can be useful for genre studies as well as automatic text type identification in Web corpora (e.g.~\newcite{asheghi2014designing}, \newcite{rehm2008towards}, and \newcite{dalan2016genre}). We plan to pursue several of these approaches in future work (see Section \secref{sec:discussion}). 

Our main contributions in this paper consist in (1) presenting a genre-balanced, richly-annotated web corpus which is more than double (and in some cases 20 times) the size of the largest similar gold annotated resources; (2) evaluating tailored NLP tools which have been trained specifically for this task and comparing them to off-the-shelf tools; and (3) demonstrating the added value of rich annotations and ensembling of multiple concurrent tools, which improve the accuracy of each task by incorporating information from other tasks and tool outputs.

In the remainder of this paper, we will describe and evaluate the quality of the dataset: In the next section, we present the data collected for our corpus, \textbf{AMALGUM} ({\bf A} {\bf M}achine-{\bf A}nnotated {\bf L}ookalike of {\bf GUM}), which covers eight English web genres for which we can obtain gold standard training data: news, interviews, travel guides, how-to guides, academic papers, fiction, biographies, and forum discussions. Section 2 describes the data and how it was collected, Section 3 presents annotations added to the corpus and the process of using information across annotation layers to improve NLP tool output, Section 4 evaluates the quality of the resulting data, and Section 5 concludes with future plans and applications for the corpus. Our corpus is available at \url{https://github.com/gucorpling/amalgum}.

\section{Data}

\subsection{Corpus Composition}

In order to ensure relatively high-quality NLP output and an open license, we base our dataset on the composition of an existing, smaller web corpus of English, called GUM (Georgetown University Multilayer corpus \cite{Zeldes2017}). The GUM corpus contains data from the same genres mentioned above, currently amounting to approximately 130,000 tokens. We use the term genre somewhat loosely here to describe any recurring combination of features which characterize groups of texts that are created under similar extralinguistic conditions and with comparable communicative intent (cf.~\cite{BiberConrad2009}). The corpus is manually annotated with a large number of layers, including document layout (headings, paragraphs, figures, etc.); multiple POS tags (Penn tags, CLAWS5, Universal POS); lemmas; sentence splits and types (e.g.~imperative, wh-question etc., \cite{LeechEtAl2003}); Universal Dependencies \cite{NivreEtAl2017}; (non-)named entity types; coreference and bridging resolution; and discourse parses using Rhetorical Structure Theory \cite{MannThompson1988}. We use this data both to train in-domain tools to create our automatic annotations (Section 3), and to evaluate the accuracy of our architecture (Section 4).

To maximize similarity with the types of data included in our corpus, we use identical or similar sources to those in GUM, with some small differences as indicated in Table \ref{tab:corpuscontents}.

\begin{table*}[bhth]
\begin{center}
\begin{tabular}{l|crl|crrl|}
 &  \multicolumn{3}{|c|}{\textbf{GUM}} & \multicolumn{4}{|c|}{\textbf{AMALGUM}}   \\
\textbf{genre} & \textbf{docs} & \textbf{tokens} & \textbf{source} & \textbf{docs} & \textbf{tokens} & \textbf{mean size} & \textbf{source} \\
\hline
\textit{academic} & 16 & 15,110 & various & 662 & 500,285 & 755 & MDPI \\
\textit{biography} & 20 & 17,951 & Wikipedia & 600 & 500,760 & 835 & Wikipedia \\
\textit{fiction} & 18 & 16,307 & various & 457 & 500,088 & 1,094 & Project Gutenberg \\
\textit{forum} & 18 & 16,286 & Reddit & 682 & 500,412 & 724 & Reddit \\
\textit{how-to} & 19 & 16,920 & wikiHow & 613 & 500,014 & 816 & wikiHow \\
\textit{interview} & 19 & 18,037 & Wikinews & 778 & 500,090 & 636 & Wikinews \\
\textit{news} & 21 & 14,094 & Wikinews & 686 & 500,600 & 733 & Wikinews \\
\textit{travel} & 17 & 14,955 & Wikivoyage & 482 & 500,680 & 1,083 & Wikivoyage \\
\hline
\textbf{total} & 148 & 129,660 & & 4,960 &  4,002,929  & 807 &  \\
 \\
\end{tabular}
\end{center}
\caption{\label{tab:corpuscontents} Corpus contents of GUM and AMALGUM. }
\end{table*}

As shown in the table, we collect equal portions of approximately 500,000 tokens from each source. This amounts to a total corpus size of about 4 million tokens, more than double the size of benchmark corpora for entity and coreference resolution (about 1.6 million tokens annotated for English coreference in OntoNotes) and 4--20 times the size of the largest English discourse treebanks (around 1M tokens for the shallow parsed PDTB \cite{PrasadWebberLeeEtAl2019}, and 200K tokens for RST-DT with full document trees \cite{CarlsonEtAl2003}, both of which are restricted to WSJ text). The close match in content to the manually annotated GUM means that we expect AMALGUM to be particularly useful for applications such as active learning when combining resources from the gold standard GUM data and the automatically prepared data presented here.

\subsection{Document Filtering}

Documents were scraped from sites that provide data under open licenses (in most cases, Creative Commons licenses). The one exception is text from Reddit forum discussions, which cannot be distributed directly. To work around this, we distribute only stand-off annotations of the Reddit text, along with a script that recovers the plain text using the Reddit API and then combines it with the annotations.

In order to ensure high quality, we applied a number of heuristics during scraping to rule out documents that are likely to differ strongly from the original GUM data: for fiction, we required the keyword ``fiction'' to appear in Project Gutenberg metadata, and disqualified documents that contained archaic forms (e.g.~``thou'') or unrestored hyphenation (e.g.~tokens like ``disre-'', from broken ``disregard'') by using a stop list of frequent items from sample documents marked for exclusion by a human. For Reddit, we ruled out documents containing too many links or e-mail addresses, which were often just lists of links. In all wiki-based genres, references, tables of contents, empty sections, and all other portions that contained non-primary or boilerplate text were removed.

\subsection{Document Extent}

Since the corpus described here is focused on discourse-level annotations, such as coreference resolution and discourse parses, we elected to only use documents similar in size to the documents in the benchmark GUM data, which generally has texts of about 500--1,000 words. This restriction was also necessary due to the disparity across genres, and especially in academic articles, which can be quite long, or most extremely in fiction, where Project Gutenberg data comprises entire books. Following the strategy used by GUM, we only select texts that have a minimum of around 400 words, and for data longer than 1,000 words, we choose contiguous sections from each source by selecting a main heading from a random part of the text, and gathering all subsequent paragraphs until 1,000 words are exceeded, but excluding document final headings if the subsequent paragraph is not included. 

For some of the genres in the corpus, additional strategies were employed: for Reddit, documents were created by randomly selecting a post that was between 25--500 words, then recursively collecting random responses until a size between 500--1,000 words was reached. For fiction, consecutive headings in the initial section were forbidden in order to avoid front matter and table of contents materials, usually meaning that a sample begins at the start of a chapter. Travel, how-to guides, news, interviews, and biographies generally begin at the top of the article after boilerplate removal, and academic data typically has one or two sections of a paper or an entire, very short paper.\footnote{An anonymous reviewer has suggested that there may be interesting differences between the genre-based subcorpora which could feed into a discussion of the notion of genre as reflected in the annotations of the corpus. Although a more developed discussion of this idea is outside the scope of the current paper, we fully agree that this is an interesting avenue to pursue and envision this as one of the applications of our corpus.}

\section{Annotation}\label{sec:anno}

Our annotated data follows the structure of the already available GUM corpus closely, meaning that for all documents we tokenize, tag, and lemmatize; add Universal Dependency parses and morphological features; add sentence types and document structure annotations (paragraphs, headings, etc.); perform entity and coreference resolution; and add full RST discourse parses. Although our automatic annotations will inevitably contain errors, our initial goal for this dataset is to obtain a resource with NLP quality that is substantially better than out-of-the-box NLP, and can therefore be used to improve tools later on. To do so, we rely on three strategies:

\begin{enumerate}
    \item Retraining tools with genre-specific data
    \item Use model stacking \cite{Wolpert1992}, applying multiple tools to a single annotation type and combining their outputs
    \item Incorporate information from other layers that is not normally available to out-of-the-box tools
\end{enumerate}

\paragraph{Structural Markup} All documents contain basic structural markup that could be reliably extracted across genres, usually from HTML tags, including headings, paragraphs, position of figures and captions, bulleted lists, speaker information (for Reddit), and highlighting via boldface or italics. This information is useful for subsequent layers, since paragraphs, headings etc.~restrict sentence boundaries, but also because the presence of markup is a predictive feature for discourse structure (see {\it Discourse Parsing} below).

\paragraph{Tokenization and Tagging} Documents are initially tokenized using a custom rule-based tokenizer with a large number of postprocessing rules based on spot-checking of common mistakes in the data. For example, the tokenizer handles typical sequences of expressions for phone numbers and opening times used in travel guides from Wikivoyage, which out-of-the-box tokenizers often split incorrectly. For POS tagging we train an ensemble model. This model takes 4 models' tag predictions as input, fits an XGBoost model \cite{ChenGuestrin2016} to them and then predicts the final tag of the tokens. The 4 models we use here are Flair's \cite{AkbikBergmannVollgraf2019} default POS tagging model which is trained on OntoNotes, Flair re-trained on GUM, StanfordNLP \cite{QiDozatZhangEtAl2018} pretrained on EWT, and StanfordNLP re-trained on GUM.\footnote{One reviewer has asked how StanfordNLP compares to other available libraries, such as Spacy (\url{https://spacy.io/}). While we do not have up to date numbers for Spacy, which was not featured in the recent CoNLL shared task on Universal Dependencies parsing, the most recent numbers reported in \cite{ZeldesSimonson2016} do not suggest that it would outperform StanfordNLP.} Since we need training data for both re-training the models and training the ensemble model, we split GUM's train set into 5 folds. Each time, we use 4 of the folds to re-train Flair and StanfordNLP and then make predictions on the remaining fold. The predictions on all these 5 folds, together with Flair OntoNotes and StanfordNLP EWT predictions, constitute the training data for the ensemble model. At test time, the four base models are run in parallel and their outputs are fed into the ensemble classifier in a pipeline. This approach improves the accuracy of the POS tagger, which in turn gets used by the machine learning sentence splitter from \newcite{YuZhuLiuEtAl2019}, which was retrained on our genres and incorporated a number of features, including POS tags, as well as the dependency parser (see below). Sentence boundaries from the sentencer splitter are also used as a basis for syntactic parsing, sentence type annotation, and maximal units for discourse parsing (see below).

\paragraph{Dependency Parsing} Universal Dependency parses and morphological features are extracted using StanfordNLP, which was retrained on our genres using the latest GUM data, and configured to use the high-accuracy tokenization and POS tags from the previous components, outperforming StanfordNLP's base accuracy (see Section \secref{sec:eval}). No ensembling was done at this stage, though we are considering the application of parser interpolation or ensembling to this task in future work \cite{KuncoroBallesterosKongEtAl2016}.

\paragraph{Coreference and Entity Resolution} Following GUM's annotation scheme, we attempt to classify all entity mentions in the corpus, including nested non-named and pronominal entities, requiring coreference resolution. Since our gold standard training data is limited in size, we rely on a combination of the knowledge-driven system, xrenner \cite{ZeldesZhang2016}, for in-vocabulary items and on neural sequence labeling using large trainable contextualized BERT embeddings \cite{devlin2018bert} for ambiguous or out of vocabulary entities. Coreference resolution is done based on linguistic features using xrenner's XGBoost implementation which is more robust than state of the art lexicalized neural coreference architectures for our data (see Section \secref{sec:eval} \    for a comparison of systems).

\paragraph{Discourse Parsing} For discourse parsing we use a modified version of the freely available DPLP parser \cite{JiEisenstein2014}, which is known to perform at or close to SOTA for English \cite{MoreyMullerAsher2017} on a number of metrics. Since we have access to additional annotation layers, we featurize structural markup (paragraphs, headings, lists, etc.) and add parse-based sentence types and the genre as features in order to boost out-of-the-box performance. The parser operates on automatically segmented discourse units generated by ToNy \cite{MullerBraudMorey2019}, a discourse segmenter trained specifically on the GUM data. The resulting RST trees are constrained to use the parsed sentence splits as maximal units: no elementary discourse unit (EDU) is permitted to be larger than one sentence, and headings, captions, and speaker ``turns'' (in Reddit) are guaranteed to be split from their environment thanks to the structural markup. We also feed our system the predicted discourse function labels from a Flair sentence classifier trained on RST-DT and out-of-the-box sentiment and subjectivity scores using TextBlob's\footnote{\url{https://textblob.readthedocs.io/en/dev/}} pretrained model as features.


\section{Evaluation}\label{sec:eval}

To evaluate the quality of our tools, we use two different test sets: the test set from the gold standard GUM corpus, which has very similar data to ours and was used for the 2018 CoNLL shared task on dependency parsing, and a random sample of contiguous AMALGUM sentences from 16 documents (two `snippets' from each genre), manually corrected on all layers and totalling some 2,000 tokens equally divided across the subcorpora. The use of the shared task test set also facilitates comparison with other tools that have been, or will be evaluated on that dataset in the future. Some example analyses from the automatically processed AMALGUM data are given in the Appendix for reference.

\paragraph{Tokenization and Tagging}

Tokenization is evaluated on the GUM test set and AMALGUM snippets. As a baseline for high-quality ``out-of-the-box'' NLP, we use StanfordNLP's default English model (pre-trained on EWT), which we compare to StanfordNLP trained on GUM and our own tokenizer (Table \ref{tab:tokenization}). Although tokenization is marginally better on GUM test for the StanfordNLP GUM model (+0.03\%), the results on the AMALGUM sample (+0.73\%), as well as qualitative inspection of some of the frequent patterns in the complete AMALGUM data, suggest that our own tokenizer is the better choice overall for this corpus.

\begin{table}[h!bt]
\centering
\begin{tabular}{lccc}
& P & R & F1 \\
\hline\hline
& \multicolumn{3}{c}{GUM test} \\
\hline
\textit{StanfordNLP (EWT)} & 98.41 & 99.97 & 99.19 \\
\textit{StanfordNLP (GUM)} & 99.93 & 99.91  & \textbf{99.92} \\
\textit{this paper}
  & 99.87 & 99.90 & 99.89 \\
\hline\hline
& \multicolumn{3}{c}{AMALGUM} \\
\hline
\textit{StanfordNLP (EWT)} & 99.35 & 99.12 & 99.23 \\
\textit{StanfordNLP (GUM)} & 99.80 & 98.51  & 99.15 \\
\textit{this paper} 
  & 99.85 & 99.90 & \textbf{99.88} \\
\end{tabular}
\caption{Tokenization Performance.}
\label{tab:tokenization}
\end{table}

For tagging, we also use StanfordNLP's pretrained model (EWT), as well as a model retrained on GUM as out-of-domain and in-domain baselines. The performance of these baselines along with the ensemble tagger (this paper) are reported in Table \ref{tab:pos-tagging}, clearly showing that in-domain training data and ensembling each boost performance substantially.

\begin{table}[hhtb]
    \centering
\begin{tabular}{lcc}
    &GUM test & AMALGUM\\
  \hline
  \hline
  \textit{StanfordNLP (EWT)} & 93.07 & 93.99 \\
  \textit{StanfordNLP (GUM)} & 95.85 & 96.97 \\
  \textit{XGBoost (4 models)}
  & \textbf{97.04} & \textbf{97.37} \\
  \hline
    \end{tabular}
        \caption{Tagging Performance (Accuracy).}
        \label{tab:pos-tagging}
\end{table}

\paragraph{Dependency Parsing}

Dependency parsing is evaluated using gold tokenized input, but predicted POS tags from the ensemble tagger described in Section \secref{sec:anno}.  We report the standard evaluation metrics as used in the CoNLL shared task on UD parsing. The baseline is again StanfordNLP's out-of-the-box performance, compared to a model trained on GUM and the same model when forced to use the ensemble's superior predicted tags (tokenization is gold in all cases for comparability). We note that these results represent a new SOTA score on GUM test as compared to the 2018 shared task (LAS 85.05, UAS 88.57, \cite{ZemanHajicPopelEtAl2018}).

\begin{table}[hhtb]
\centering
\begin{tabular}{lcc}
& UAS & LAS \\\hline\hline
& \multicolumn{2}{c}{GUM test} \\
\hline
\textit{StanfordNLP (EWT)} & 86.89 & 83.66 \\
\textit{StanfordNLP (GUM)} & 89.32 & 85.49 \\
\textit{this paper}     & \textbf{89.47} & \textbf{85.89} \\
\hline\hline
& \multicolumn{2}{c}{AMALGUM} \\
\hline
\textit{StanfordNLP (EWT)} & 85.07 & 81.41 \\ 
\textit{StanfordNLP (GUM)} & 87.76 & 84.38 \\
\textit{this paper}     & \textbf{88.81} & \textbf{85.77} \\

\end{tabular}
\caption{Dependency Parsing Performance.}
\label{tab:dep}
\end{table}

\paragraph{Coreference and Entity Resolution}


A key challenge in using both SOTA coreference resolution and entity recognition approaches is that the two tasks are interdependent: if our coreference resolution system is not optimal for entity recognition, we cannot simply replace it with a separate entity recognizer/classifier, since we will obtain incoherent coreference chains whenever output mention spans from the coreferencer do not match the entity recognizer. In other words, we need to decide what the entity spans are and force the coreference resolution architecture to use those as candidates, or the annotations will not line up. This is complicated by the fact that many coreference resolution systems trained on OntoNotes, which does not include singleton mentions (entities mentioned only once), also do not output singletons, and therefore cannot be used to exhaustively determine spans for entity classification. To tackle this challenge, we evaluate separate coreference and entity resolution solutions, as well as combinations of multiple tools to achieve the best possible solution which maintains a coherent analysis.

We first evaluate the performance of xrenner \cite{ZeldesZhang2016}, which performs both coreference resolution and nested entity recognition jointly, including singletons (i.e. non-coreferring entity mentions). Since the CoNLL-2012 English dataset excludes many common coreference types such as indefinite anaphors, compound modifiers, and copular predicates, we must also retrain the SOTA system for comparison on GUM, which targets unrestricted coreference and contains all these phenomena. We select the coarse-to-fine, end-to-end neural coreference model \cite{lee-etal-2018-higher} with BERT-base embeddings \cite{Joshi2019BERTFC} as the SOTA baseline system. For consistency, we remove singleton mentions in the evaluation, following the standard CoNLL metrics, though our final output will contain singletons to match the goals and annotation scheme taken over from GUM.

Since the training data for GUM is limited, \newcite{Joshi2019BERTFC}'s model does not achieve satisfying scores out-of-the-box on GUM test as in the CoNLL dataset. Table \ref{tab:coreferenceresults} shows that xrenner outperforms the SOTA coref system by 10.0\% on average F1 on the GUM test set.\footnote{F1 scores reported throughout the paper are micro-averaged.} For AMALGUM snippets, xrenner achieves 79.3\% on average F1, which is 14.9\% higher than \newcite{Joshi2019BERTFC}'s model. However in the highest scoring xrenner model, it turns out that entity types for many \textit{non-coreferential} mentions are more often incorrect than for the state of the art nested entity recognition system \cite{shibuya_nested_2019}, evaluated below. We therefore considered using that system for mention detection and classification, and report slightly lower scores using this combination of entity recognition and coreference resolution (xrenner + S\&H in the table).

\begin{table*}[h!bt]
    \centering\small
    \begin{tabular}{lcccccccccc}
    \hline
     & \multicolumn{3}{c}{MUC} & \multicolumn{3}{c}{B$^3$} & \multicolumn{3}{c}{CEAF$_{\phi4}$} & Avg. \\
     & P & R & F1 & P & R & F1 & P & R & F1 & F1 \\
     \hline\hline
    \multicolumn{11}{c}{GUM test}\\\hline

\textit{    \newcite{Joshi2019BERTFC} }& 74.4 & 48.3 & 58.6 & 66.9 & 26.0 & 37.5 & 60.5 & 11.2 & 18.9 & 41.4 \\
    \textit{xrenner }& 53.4 & 69.5 & 60.4 & 41.1 & 61.3 & 49.2 & 39.9 & 50.2 & 44.5 & \textbf{51.4} \\
       \textit{ xrenner + S\&H types} & 53.2 & 69.6 & 60.3 & 40.5 & 61.6 & 48.9 & 39.7 & 50.4 & 44.4 & 51.2 \\\hline\hline
    \multicolumn{11}{c}{AMALGUM}\\\hline
    \textit{\newcite{Joshi2019BERTFC}} & 89.2 & 59.6 & 71.5 & 86.0 & 51.5 & 64.4 & 83.2 & 43.8 & 57.4 & 64.4 \\
    \textit{xrenner} & 89.1 & 79.0 & 83.7 & 83.8 & 74.3 & 78.7 & 81.0 & 70.7 & 75.5 & \textbf{79.3} \\
        \textit{xrenner + S\&H types }& 88.5 & 77.5 & 82.6 & 83.1 & 72.4 & 77.4 & 81.6 & 68.0 & 74.2 & 78.1 \\\hline\hline
    \end{tabular}
    \caption{Coreference Resolution Performance. The average F1 of MUC, B$^3$ and CEAF$_{\phi4}$ is the main evaluation metric.}
    \label{tab:coreferenceresults}
\end{table*}

As nested entity recognition is itself an essential component for coreference resolution, we experimented with a number of configurations and chose the implementation in  \newcite{shibuya_nested_2019}, which uses a first/second/n-best path CRF sequence decoding model with BERT-large embeddings, as it achieves higher F1 on entity type recognition than the out-of-the-box xrenner model, as shown in Table \ref{tab:NNERresults}. 

\begin{figure*}[b!]
    \centering
    \includegraphics[width=0.95\textwidth]{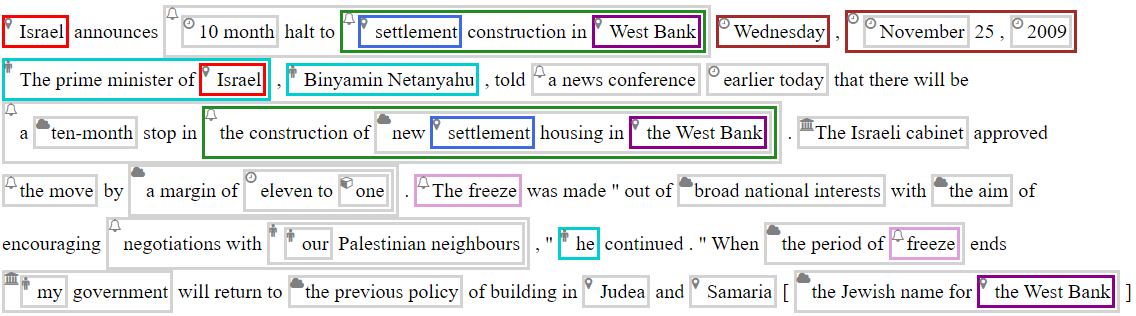}
    \caption{Xrenner's coreference and entity predictions on an AMALGUM news snippet.  Coreferent mentions are colored (e.g. \cfbox{red}{Israel} and \cfbox{cyan}{the prime minister of Israel} are boxed in red and cyan), and entity types are indicated by icons: \textsc{place} \faMapMarker, \textsc{person} \faMale, \textsc{time} \faClockO, \textsc{event} \faBellO, \textsc{abstract} \faCloud, and \textsc{object} \faCube.}
    \label{fig:corefpred}
\end{figure*}

In order to capitalize on both xrenner's superior coreference resolution performance and the predictions of the model from \newcite{shibuya_nested_2019}, we created a hybrid model by injecting \newcite{shibuya_nested_2019}'s prediction of entity types on identical token spans recognized by the coreferencer (xrenner), which were then fed to the modified coreference resolution system (xrenner + S\&H types). This improved F1 by 1.36\% on GUM test and 5.16\% on AMALGUM snippets. Figure \ref{fig:corefpred} visualizes the coreference and entity predicted by xrenner on a sample document in AMALGUM, giving an idea of the quality of predicted analyses.

The modified xrenner significantly outperforms \newcite{shibuya_nested_2019} on AMALGUM and only scores 1.78\% lower on GUM test. We decided to adopt this as the final coreference and entity resolution model for AMALGUM, despite the slightly lower coreference scores for xrenner + S\&H in Table \ref{tab:coreferenceresults}, since coherent mention chains are essential for research and applications of discourse annotation, in which  entity spans and types should not contradict the entity predictions on the coreference layer, and entity types should match throughout a coreference chain.

\begin{table}[h!bt]
    \centering\small
    \begin{tabular}{lccc}
    \hline
     & P & R & F1  \\
     \hline\hline
    \multicolumn{4}{c}{GUM test}\\\hline
  \textit{  \newcite{shibuya_nested_2019}} & 65.16 & 63.67 &  \textbf{64.41}   \\
   \textit{ xrenner} & 58.39 & 64.45 &  61.27 \\
   \textit{ xrenner + S\&H types} &   59.70 & 65.87 & 62.63 \\\hline\hline
    \multicolumn{4}{c}{AMALGUM}\\\hline
   \textit{ \newcite{shibuya_nested_2019}} & 77.58 & 60.03 & 67.69  \\
   \textit{ xrenner} &  69.77 & 63.35 &  66.40 \\
   \textit{ xrenner}  + S\&H types & 75.12 & 68.33 & \textbf{71.56}   \\\hline
    \end{tabular}
    \caption{Nested Entity Recognition Performance.}
    \label{tab:NNERresults}
\end{table}

\paragraph{Discourse Parsing} 

For discourse parsing, we compare our results to the default system settings of DPLP, which uses dependency parses, lexical features and the graph topology conveyed by the parser states themselves, but no information on markup such as paragraphs and headings, or genre and other high level features (\textit{J\&E14} baseline). Our own tagging and parsing preprocessor uses markup information, unit predictions from ToNy, and genre and sentence types (declarative, imperative, etc.) from other layers as features, next to the NLP inputs described in Section \secref{sec:anno}. For comparability, both systems were given gold discourse unit segmentation for the results in the table, as is standard in discourse parsing evaluation, for both GUM test and the AMALGUM snippets.


\begin{table}[hbt]
\centering
\begin{tabular}{lccc}
& span & nuclearity & relation \\
\hline\hline
& \multicolumn{3}{c}{GUM test} \\
\hline
\textit{J\&E14} & 67.62 & 43.94 & 24.17 \\
\textit{J\&E14+multilayer}
& \textbf{77.98} & \textbf{61.79} & \textbf{44.07} \\
\hline\hline
& \multicolumn{3}{c}{AMALGUM} \\
\hline
\textit{J\&E14} & 73.93 & 46.68 & 25.06 \\
\textit{J\&E14+multilayer}
& \textbf{84.03} & \textbf{65.01} & \textbf{45.13} \\
\end{tabular}
\caption{RST Discourse Parsing Results.}
\label{tab:rsteval}
\end{table}

The results in Table \ref{tab:rsteval} indicate that the added multilayer annotations improve accuracy greatly, and that as in the case for coreference resolution, the AMALGUM snippets are less challenging than the full length documents from GUM test. Two possible reasons for this include the fact that document and section beginnings typically contain headings and figures, which are usually labeled as \textsc{preparation} and \textsc{background} and attached to the right, and the fact that smaller documents have less complex non-terminal structures resulting from relations between paragraphs, which are substantially more difficult to classify (see the gold and predicted samples in the Appendix for an example of these structures). Realistic, full parse accuracy is therefore likely to be closer to the GUM test set scores for full length documents, and improving this accuracy will be a main focus of our future work.

\section{Discussion and Outlook}\label{sec:discussion}

The evaluation shows the marked difference between using out-of-the-box NLP tools, 
retraining tools on within-domain data, and the application of model stacking and multilayer features across annotation tasks in producing the most accurate automatically annotated corpus data possible. In the case of the corpus presented here, it has been possible to construct a substantially larger web corpus based on a smaller gold standard dataset with some very high quality results, especially for tokenization, part of speech tagging, and to a large extent, syntactic parsing. Tagging accuracy in particular varied from just under 94\% (out-of-the-box) to around 96\% (in-domain) and a maximum of 97.37\% accuracy using model stacking, a result very close to gold standard corpora which then benefits parsing and subsequent annotations building on the tagged data.

At the same time, the accuracy achieved for discourse level annotations demonstrates that NLP tools have a long way to go before truly high quality output can be expected. Realistic accuracy on new domains for tasks such as complete coreference resolution or nested entity recognition are well below SOTA reported scores, which are possible when evaluating on OntoNotes, but less representative of what can be achieved on web data `in the wild.' Especially in the case of discourse parsing, even with in-domain training data and information from other annotation layers, results are still modest, revealing the substantial challenge in producing tools that can be applied to unseen data. The results also reveal just how indispensable additional resources beyond input text can be, such as document markup, genre, and features tagged based on other training sets (e.g.~using discourse relation labels predicted by a model trained on RST-DT). Auxiliary information from external data sets was also used to boost tagging accuracy by incorporating predicted tags from models trained on the English Web Treebank and OntoNotes, suggesting that ensembles trained on heterogeneous data sets can improve NLP results in practice.

In future work we plan to enhance the quality of the corpus using some of the techniques mentioned in Section \secref{sec:intro}, such as active learning, bootstrapping, and targeted use of crowd sourcing, with the aim of especially improving discourse level annotations, such as coreference resolution and discourse parsing. By being `better than (out-of-the-box) NLP,' we hope this corpus will be both a target and a resource for research on genre variation and discourse level phenomena, as well as the development of increasingly accurate tools.

\section{Bibliographical References}
\label{main:ref}

\bibliographystyle{lrec}
\bibliography{main}

\section{Appendix: Sample Analyses}

The figures below visualize predicted output for discourse parsing for one of the snippets from the AMALGUM test set. A reference discourse parse is provided as well. For coreference and entity resolution, see Figure \ref{fig:corefpred}.

\begin{figure*}[b!]
    \centering
    \includegraphics[width=1.0\textwidth]{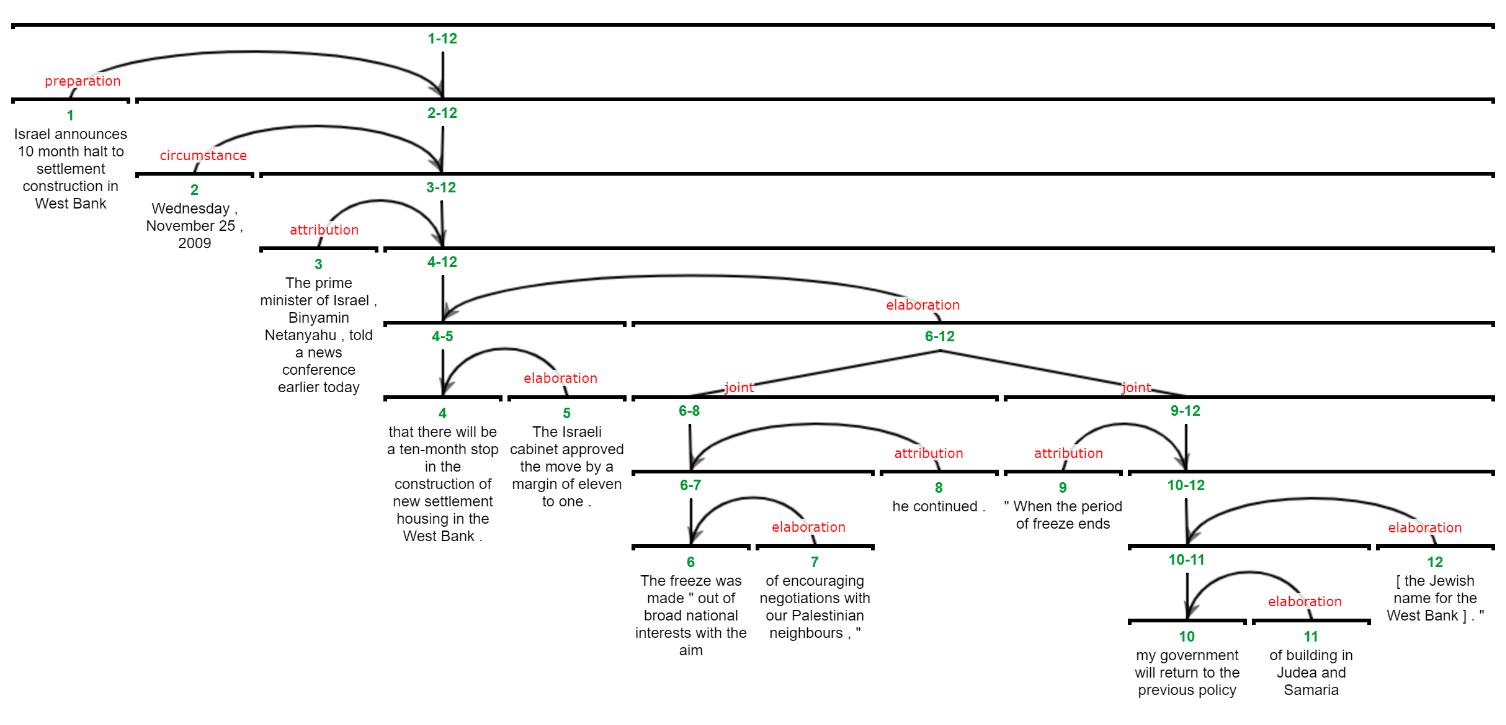}
    \caption{Predicted discourse parse for the same news text; errors compared to Figure \ref{fig:rstgold} include viewing the circumstance clause `when the freeze ends' as \textsc{attribution} and an incorrect attachment of the \textsc{elaboration} about the name of the West Bank. In contrast to the human analyst, the parser also groups units [6-12] as a \textsc{joint}, which is not implausible.}
    \label{fig:rstpred}
\end{figure*}
\begin{figure*}[h!]
    \centering
    \includegraphics[width=1.0\textwidth]{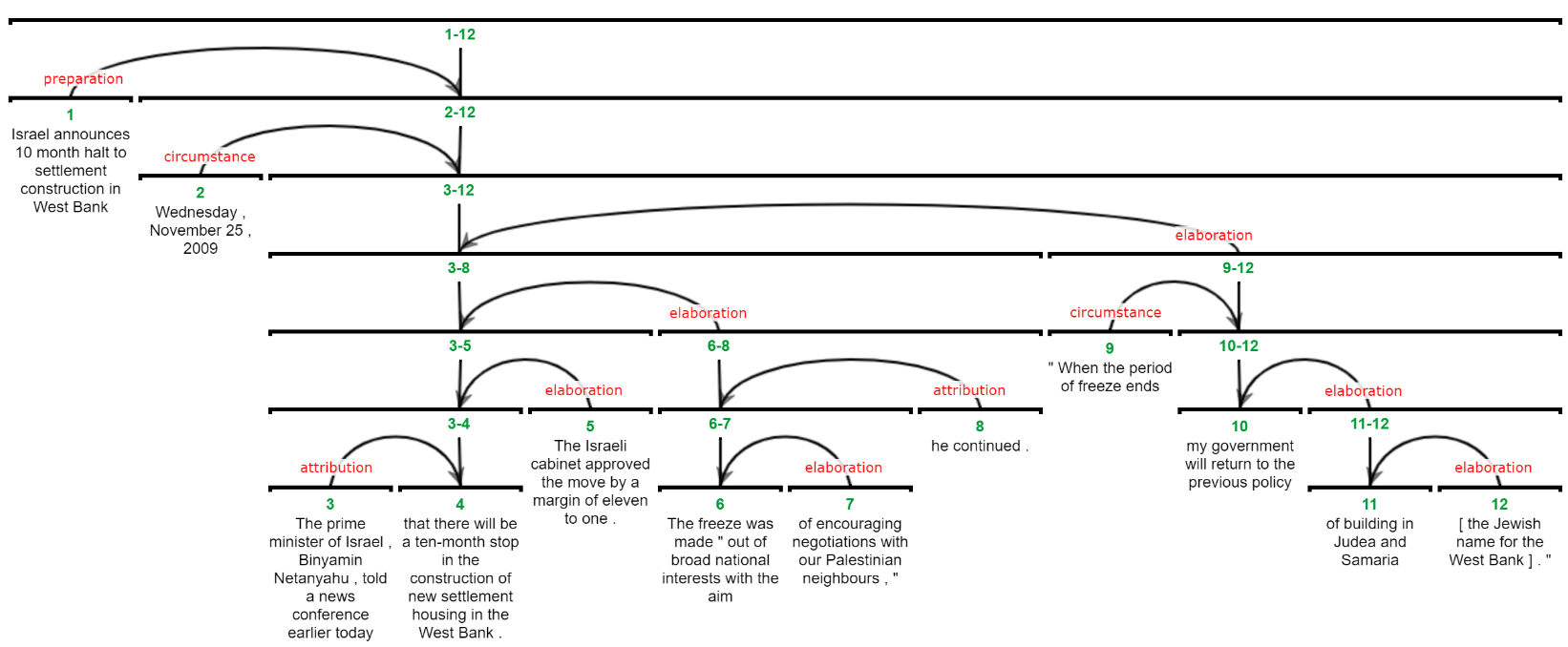}
    \caption{Manual discourse parse of the news text excerpt from Figure \ref{fig:rstpred}.}
    \label{fig:rstgold}
\end{figure*}


\end{document}